\documentclass[lettersize,journal]{IEEEtran}
\usepackage{amsmath,amsfonts}
\usepackage{algorithmic}
\usepackage{array}
\usepackage{textcomp}
\usepackage{stfloats}
\usepackage{url}
\usepackage{verbatim}
\usepackage{graphicx}
\usepackage{multirow}
\usepackage{caption}
\usepackage{subcaption}
\def\BibTeX{{\rm B\kern-.05em{\sc i\kern-.025em b}\kern-.08em
    T\kern-.1667em\lower.7ex\hbox{E}\kern-.125emX}}
\usepackage{balance}
\begin{document}
\title{Experimental Assessment of a Forward-Collision Warning System Fusing Deep Learning and Decentralized Radio Sensing}

\author{
	Jorge D. C\'ardenas,~\IEEEmembership{Student Member, IEEE}, 
	Omar Contreras-Ponce,\\%
	Carlos A. Guti\'errez,~\IEEEmembership{Senior Member, IEEE}, 
	Ruth Aguilar-Ponce,~\IEEEmembership{Member, IEEE},\\%
	Francisco R. Castillo-Soria,~\IEEEmembership{Member, IEEE}, 
	Cesar A. Azurdia-Meza,~\IEEEmembership{Member, IEEE}
	\thanks{Manuscript submitted May 28, 2023.}
	\thanks{
		J. D. C\'ardenas-Amaya, C. A. Guti\'errez. R. Aguilar-Ponce 
		and F. R. Castillo-Soria are with the Faculty of Science, Universidad Aut\'onoma de San Luis Potos\'i, 
		San Luis Potosi 78295, Mexico (e-mail: j.cardenas@ieee.org, cagutierrez@ieee.org, ruth.aguilar@ieee.org, ruben.soria@uaslp.mx).
	}
	\thanks{O. Contreras-Ponce is with Robert Bosch M\'exico, 
		Hermosillo, México (e-mail: ponce.omar.dm@gmail.com)
	}
	\thanks{C. A. Azurdia-Meza is with the Department of Electrical Engineering, Universidad de Chile, Santiago, Chile (e-mail: cesarazurdia@uchile.cl)
	}
}
	

\markboth{Journal of \LaTeX\ Class Files,~Vol.~18, No.~9, September~2020}%
{How to Use the IEEEtran \LaTeX \ Templates}

\maketitle

\begin{abstract}
	This paper presents the idea of an automatic forward-collision warning system based on a decentralized radio sensing (RS) approach. In this framework, a vehicle in receiving mode employs a continuous waveform (CW) transmitted by a second vehicle as a probe signal to detect oncoming vehicles and warn the driver of a potential forward collision. Such a CW can easily be incorporated as a pilot signal within the data frame of current multicarrier vehicular communication systems. Detection of oncoming vehicles is performed by a deep learning (DL) module that analyzes the features of the Doppler signature imprinted on the CW probe signal by a rapidly approaching vehicle. This decentralized CW RS approach was assessed experimentally using data collected by a series of field trials conducted in a two-lanes high-speed highway. Detection performance was evaluated for two different DL models: a long short-term memory network and a convolutional neural network. The obtained results demonstrate the feasibility of the envisioned forward-collision warning system based on the fusion of DL and decentralized CW RS.

\end{abstract}


\section{Introduction}
\IEEEPARstart{F}{orward} collisions between vehicles are road accidents that cost many lives each year. Such accidents are caused mainly by driver distractions and external factors affecting the driver's capacity to perceive other vehicles, such as weather and traffic conditions \cite{WHO2022}. This problem has prompted the development of driving assistance systems that employ a variety of sensors for automatic detection and alerting of vehicles whose proximity and trajectory could pose a collision threat. The sensors employed for such applications are primarily based on ultrasonic \cite{Krishnan2018}, infrared \cite{Mita2019}, video \cite{Chen2020,Wang2018a}, and laser \cite{Zhang2018a} technology. These sensing technologies are well understood and already incorporated in modern smart vehicles. Nonetheless, further research is still required to cover the limitations inherent to each type of sensors. For example, ultrasonic and infrared sensors are affected by weather factors like fog, snow, and rain. On the other hand, video sensors depend on good lighting, whereas laser-based sensors rely on a precise pointing to target.

Radio sensing (RS) is emerging as an option to complement the capacities of the aforementioned sensors. This technology leverages on the mechanisms of reflection, scattering and Doppler dispersion of radio-frequency (RF) signals for target detection, i.e., for the detection of nearby vehicles. Environmental conditions and the lack of a line-of-sight (LOS) with the target do not represent major limitations for RS systems. Moreover, the interest in RS has increased in recent years due to the potential of this technology for seamless integration with vehicular communication systems, through a design paradigm known as joint RS and communications (JRSAC) \cite{Zhang2021}. In this paradigm, the sensing process and the transmission of data between vehicles are simultaneously performed by means of a common RF signal. This can be accomplished because the hardware resources required for both applications are similar and can be shared to transmit and receive a single signal that serves both purposes. 

Radars are the prevailing form of RS. In fact, some forward collision warning systems based on radar technology have already been developed and are in the verge of market deployment. However, the integration of radars into the JRSAC framework is a challenging research problem, mainly because the waveforms and signal frame structure of radars are not tailored for data transport. An increasing number of research projects are currently underway with the aim of finding efficient solutions to this problem, e.g., see \cite{Daniels2018,Cong2023}. The integration of radars within the JRSAC framework is further complicated by the centralized approach of these sensing systems, whereby each vehicle independently performs the transmission, reception, and processing of its own radar signal. Such a centralized approach entails full-duplex transmission capacities. This is in contrast with the characteristics of vehicular communication systems (VCS) based on current dedicated short-range communications (DSRC) \cite{IEEE802.11p} and cellular vehicle-to-everything (C-V2X)  \cite{Soto2022} standards, whose specifications consider half-duplex (simplex or dual-simplex) transmissions. 

In this paper, we explore an alternative for automatic forward collision alerting that follows a decentralized RS approach, where a vehicle in receiving mode employs an RF signal transmitted by a second vehicle as a probe to detect other oncoming vehicles. The probe signal is a continuous waveform (CW) that can be easily appended to the data signal in a process analogous to the transmission of pilot subcarriers within the data frame of modern multicarrier communication systems. Oncoming vehicles are detected by the receiver ($R_X$) based on the Doppler signature that a rapidly approaching vehicle imprints on the CW signal. A deep-learning (DL) module is employed for the timely recognition of such Doppler signatures. We assessed the feasibility of such a decentralized CW RS approach by a series of field trials conducted in a high-speed highway near the city of San Luis Potos\'i, M\'exico. The aim of these experiments was to collect realistic information about the Doppler signatures produced by vehicles approaching the $R_X$ on a potential collision path. The aim was also to gather an empirical data set comprising a variety of Doppler signatures produced not only by oncoming vehicles but also by other relevant events. Using this data set, we evaluated the detection performance of our decentralized CW RS system by considering two different DL models: a long short-term memory (LSTM) network and a convolutional neural network (CNN). To the best of the authors' knowledge, this is the first experimental demonstration of a forward collision warning system based on the fusion of DL and a decentralized CW RS approach. 

The remainder of this paper is organized as follows. The RS scenario that motivates this research work is described in detail in Section \ref{CADA}. Section \ref{TSD} provides an overview of our measurement platform and experimental protocol. The empirical data collected during our measurement experiments are described in Sections \ref{MS} and \ref{database}. Section \ref{DL} shows the architecture of the DL algorithms and the data set generated by the experiments. In Section \ref{results} we present and discuss the results of the system performance evaluation. Finally, we give our conclusions in Section \ref{conclusion}.

\section{Forward-collision warnings based on decentralized CW RS and DL}\label{CADA} 

\begin{figure}[t]
	\centering
	\includegraphics[width=0.5\textwidth]{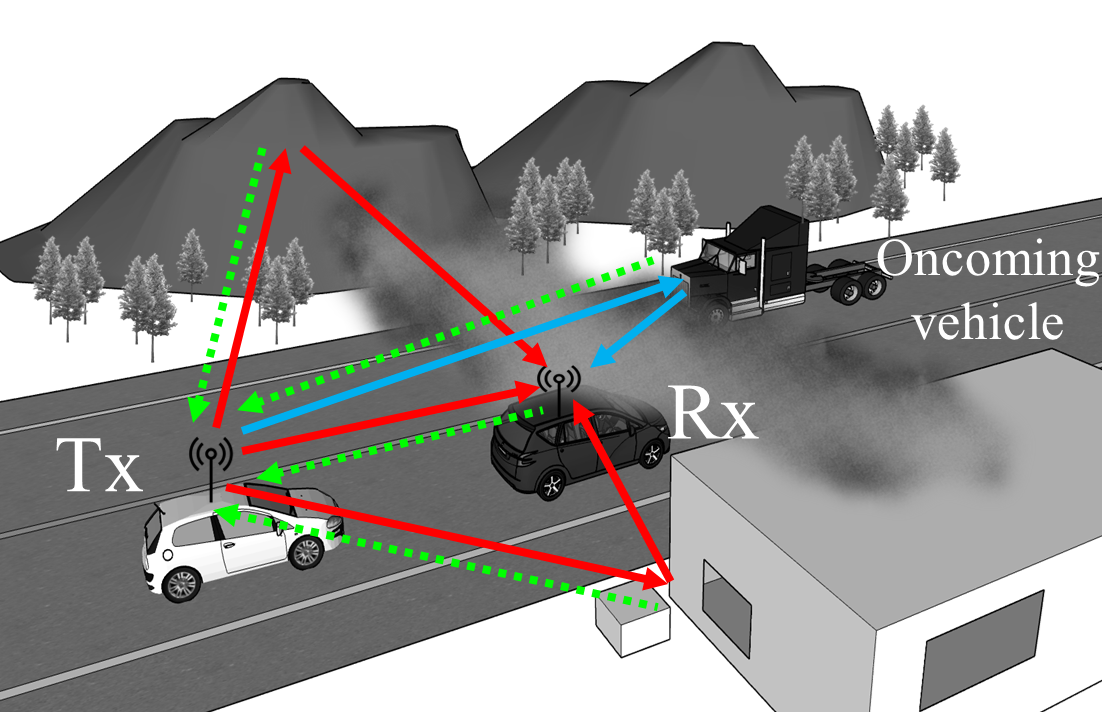}
	\caption{Centralized approach for a JRSAC system, where the solid blue and red lines correspond to the communication link and the dotted green lines represent the reflected component of the signal for the RS.}
	\label{FIG:CS}
\end{figure}

The focus of this research work is on the automatic alerting of a potential forward collision under the conditions illustrated in Fig.~\ref{FIG:CS}. In particular, we consider a RS scenario where two vehicles are driving in the same direction, one behind the other. The rear vehicle transmits a large number of data frames on an uninterrupted basis to the vehicle in front, as in the case of video streaming. The communications between the transmitter ($T_X$) and $R_X$ are assumed to follow a half-duplex vehicular communications standard, implying that only one vehicle can transmit at a time. The two vehicles are approached from ahead by a third vehicle driving on an adjacent lane, in a trajectory that poses a potential risk of frontal collision for the $R_X$. In the centralized JRSAC framework, each vehicle relies on the transmission of its own RF signal for threat detection. Hence, in the scenario illustrated in Fig.~\ref{FIG:CS}, the $T_X$ would be in capacity of detecting the oncoming vehicle, but the $R_X$ would be unable to do it, as it should remain silent during the reception of data of the $T_X$. The situation can be even more critical for the $R_X$ if its visibility toward the oncoming vehicle is compromised, e.g., by a fog bank, as depicted in Fig.~\ref{FIG:CS}. The $R_X$ could of course maintain its sensing capacities, for example, if it switches from a JRSAC mode to a disjoint mode in which the transmission of a pure sensing signal is activated at a frequency band different from that of the communications signal. However, this solution requires additional hardware resources and is not in line with the premise of using the same signal for data transmission and sensing. 

\begin{figure}
	\centering
	\includegraphics[width=0.45\textwidth]{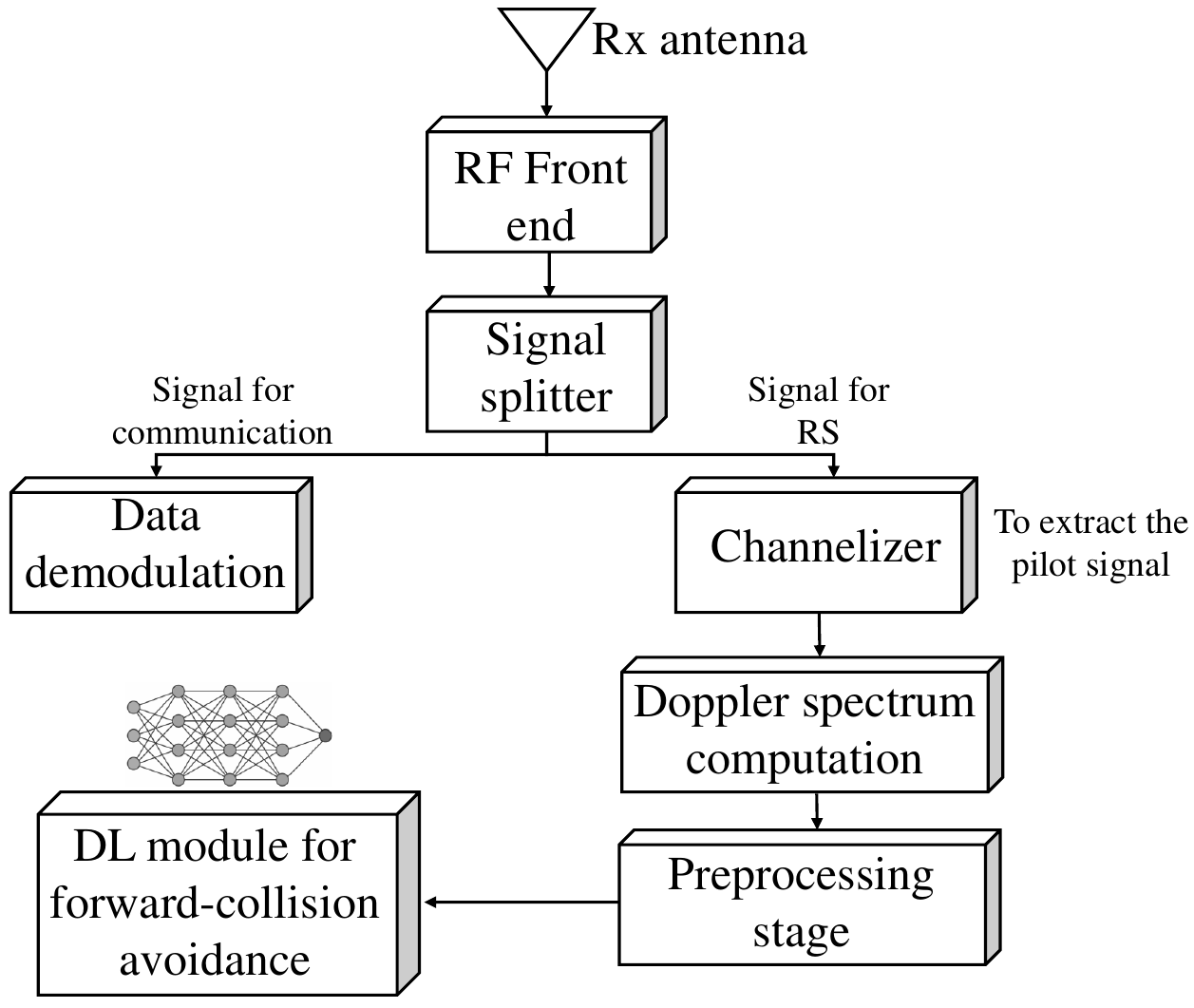}
	\caption{The received signal flow diagram of the forward-collision warning system.}
	\label{FIG:PS}
\end{figure}

An alternative that does not depend on additional hardware resources is to extract information about events happening around the $R_X$ directly from the data signal of the $T_X$. This can be accomplished, e.g., by analyzing the spectral features of the signal reflected toward the $R_X$ by the oncoming vehicle (the trajectory of this signal is indicated by the blue arrows in Fig.~\ref{FIG:CS}). Such a decentralized RS approach does not require re-engineering of the multicarrier waveforms of current VCS. This is because the spectral signature of the oncoming vehicles can be detected by computing and analyzing the Doppler spectrogram of the pilot signals embedded in these waveforms. The features of the Doppler signature imprinted by the oncoming vehicle on the spectrogram of any of such pilot signals can be characterized and classified with the aid of modern DL techniques. 
A flow diagram of the received signal processing required to implement this idea is shown in Fig. \ref{FIG:PS}. This detection strategy is similar to that described in \cite{Cardenas2023} for the detection of human falls employing WiFi signals for RS. However, the feasibility of this idea for decentralized RS in vehicular environments has yet to be verified. This paper aims to provide insight in that regard by employing empirical data to train and assess the performance of a DL-based system for the automatic detection of oncoming vehicles. The first step toward this goal is the gathering of empirical data to obtain a realistic description of the Doppler signatures of oncoming vehicles. Such a task is described in the following section.

\section{The experimental setup}\label{TSD}
\subsection{The measurement platform}

For our measurement experiments, we employed a CW channel sounding platform as the one described in \cite{Gomez-Vega2022} to mimic the transmission and reception of an unmodulated pilot signal. The platform is comprised of a RF signal generator that is carried by the $T_X$ and a spectrum analyzer mounted on the $R_X$. We conducted our experiments by considering two different frequencies for the CW probe signals: one frequency at 760~MHz, and the other at 2.5~GHz. The former frequency is reserved in some countries, such as Japan, for VCS applications. The latter frequency is in the range of the IEEE 802.11g standard for WiFi transmissions. 
We did not include measurements at the frequency band of the DSRC standards, i.e., at 5.9~GHz, because this band is beyond reach to the measurement equipment that was available for our experiments. Table \ref{tblSetting} provides an overview of the values chosen to configure the RF generator and spectrum analyzer for the two frequencies of our experiments.	The $T_X$ and the $R_X$ were also equipped with video recording and global positioning system (GPS) modules. These components are necessary to extract information of velocity, position, and for identification of the events that occurred around the $R_X$ during the experiments.

\subsection{The measurement scenario}

\begin{figure}[t]
	\centering
	\includegraphics[width=0.45\textwidth]{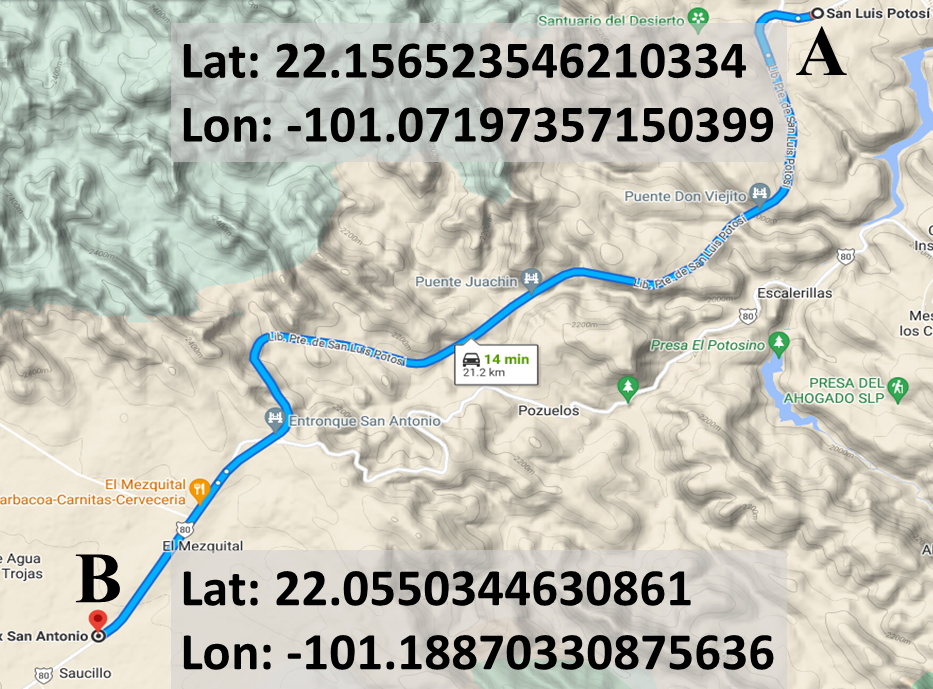}
	\caption{The measurements route.}
	\label{FIG:Map}
\end{figure}

The location chosen for our field trials was a high-speed highway near the city of San Luis Potos\'i, M\'exico. Specifically, our measurement route was a 21 km stretch of the Mexican federal highway 80D, as shown in Fig.~\ref{FIG:Map}. This highway presents the characteristics of the RS scenario described in the previous section: a high-speed two-lanes wide road that passes through a mountainous chain \cite{Gutierrez2023}. Data was recorded along a measurement circuit between points A and B in Fig.~\ref{FIG:Map}. This circuit takes 30 to 35 minutes to complete, depending on traffic conditions. The $T_X$ remained behind the $R_X$ at all times during the experiments, maintaining a separation distance with the $R_X$ between 100 and 250\,m. The speed of these vehicles ranged from 80 to 120 km/h. For each of the two frequencies (760~MHz and 2.5~GHz) of the CW signal, a data set was recorded uninterruptedly along a route comprising two circuits. 

\begin{table}
	\begin{center}
		\caption{Overview of the measurement equipment configurations.}
		\label{tblSetting}
		\begin{tabular}{| c | c | c |}
			\hline
			\textbf{Equipment} & \textbf{Parameter} & \textbf{Value} \\
			\hline
			\multirow{3}{*}{Transmitter ($T_X$)}& \multirow{2}{*}{Carrier frequency} & 760~MHz \\
			& & 2.5~GHz \\
			\cline{2-3}
			& Output level & 20~dBm \\
			\cline{1-3}
			\multirow{7}{*}{Receiver ($R_X$)}& \multirow{2}{*}{Central frequency} & 760~MHz \\
			& & 2.5~GHz \\
			\cline{2-3}
			& \multirow{2}{*}{Frequency span} & 1.5~kHz \\
			& & 2~kHz \\
			\cline{2-3}
			& Sample points & 1001 \\
			\cline{2-3}
			& Resolution bandwidth & 10~Hz \\
			\cline{2-3}
			& Sweep time & 0.40~s \\
			\cline{2-3}
			& Attenuation & 0~dB \\
			\cline{2-3}
			& Pre-amp & Off \\
			\hline
		\end{tabular}
	\end{center}
\end{table}

\section{Empirical characterization of the Doppler signature of an oncoming vehicle}\label{MS}

\begin{figure}[t]
	\centering
	\begin{subfigure}[h]{0.45\textwidth}
		\centering
		\includegraphics[width=\textwidth]{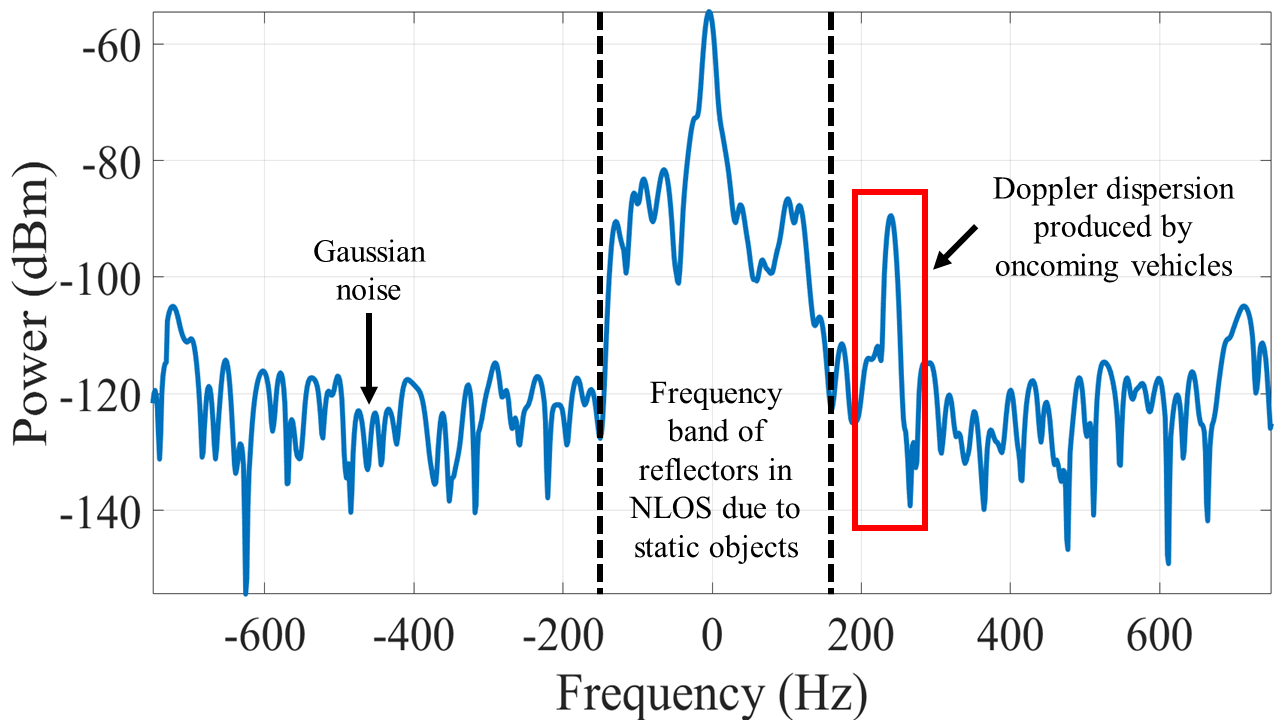}
		\caption{Snapshot of a spectrum at 760~MHz.}
		\label{FIG:signal760}
	\end{subfigure}
	\hfill
	\begin{subfigure}[h]{0.45\textwidth}
		\centering
		\includegraphics[width=\textwidth]{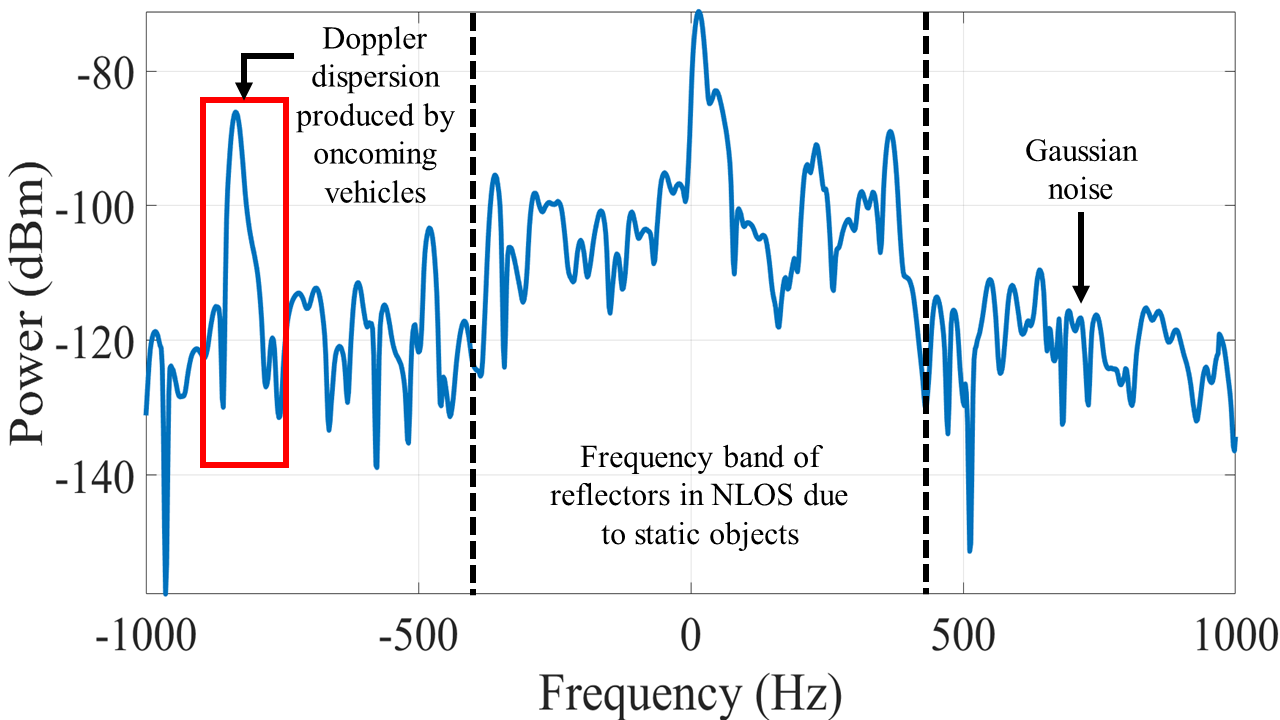}
		\caption{Snapshot of a spectrum at 2.5~GHz.}
		\label{FIG:signal2500}			
	\end{subfigure}
	\caption{Examples of the Doppler spectra measured in presence of an oncoming vehicle.}
	\label{FIG:SpectrumsOriginal}
\end{figure}

With the settings of Table \ref{tblSetting}, our measurement platform was able to record almost three snapshots per second of the received CW signal's Doppler spectrum. Figure~\ref{FIG:SpectrumsOriginal} shows two snapshots of the Doppler spectra recorded at 760~MHz (Fig.~\ref{FIG:signal760}) and 2.5~GHz (Fig.~\ref{FIG:signal2500}) during time instants at which the $R_X$ was being approached by an oncoming vehicle. These two spectra illustrate the typical frequency dispersion effects produced 
during the mobile reception of a CW signal. Specifically, we can identify the LOS component of the received signal as the spectral spike of the highest power, which is located around the origin. 
In addition, we can observe the presence of other high-power elements within a frequency band centered at the origin and bounded by a frequency of $\pm169$~Hz for the spectrum at 760~MHz and $\pm402$~Hz for the spectrum at 2.5~GHz. The spectral elements in this band correspond to the non-LOS (NLOS) components of the received signal arriving at the $R_X$ by the mechanisms of reflection and scattering from static objects around the road. The boundaries of this band of NLOS power can be determined following the well-known formula of the maximum Doppler shift due to reflection/scattering by static objects
\begin{align}
\label{DopplerMax}
\Delta f_{\max} &= \frac{v_T(t)+v_R(t)}{c} f_c
\end{align}
where $v_T(t)$ and $v_R(t)$ are the instantaneous speeds of the $T_X$ and the $R_X$ respectively, $c$ is the speed of light and $f_c$ is the carrier frequency \cite{Gutierrez2023}. For frequencies outside the interval $[-\Delta f_{\max}, \Delta f_{\max}]$, the measured Doppler spectra comprise mostly noise. However, the examples shown in Fig.~\ref{FIG:SpectrumsOriginal} show an additional spectral spike above the noise level and located at a frequency outside of $[-\Delta f_{\max}, \Delta f_{\max}]$. This term is caused by the reflection of the CW signal from the oncoming vehicle. In fact, we can infer that the oncoming vehicle is approaching the $R_X$ if the frequency of such a spectral spike is larger than $\Delta f_{\max}$, as in the case shown in Fig.~\ref{FIG:signal760}. In turn, we can infer that the oncoming vehicle has already passed by the $R_X$ and is driving away if this spectral spike has a negative frequency smaller than $\Delta f_{\max}$, as in Fig.~\ref{FIG:signal2500}.  

\begin{figure}[t]
	\centering
	\includegraphics[width=0.5\textwidth]{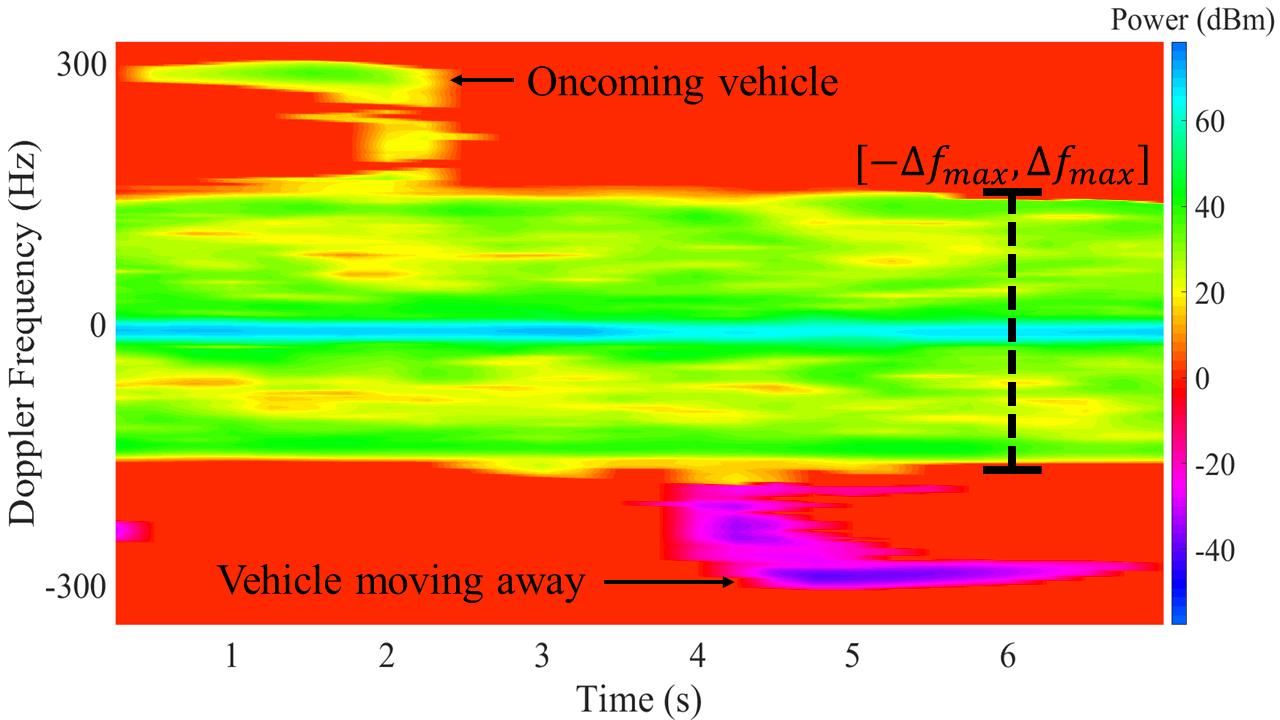}
	\caption{Measured spectrogram showing the Doppler signature of an event involving an oncoming vehicle.}
	\label{FIG:CompleteSeq}
\end{figure}

Figure \ref{FIG:CompleteSeq} shows a complete sequence of contiguous Doppler spectra recorded during an event involving an oncoming vehicle. The spectral signature shown in this figure conveys all the information required by the $R_X$ to detect the oncoming vehicle. Nonetheless, only the first part of the signature is relevant for the detection of the approaching vehicle, i.e., the part where the frequency of the spectral spike observed outside of $[-\Delta f_{\max}, \Delta f_{\max}]$ is positive and nearly constant. The middle part, where an abrupt drift from a positive frequency to a negative frequency is observed, is not helpful because this drifting effect indicates that the two vehicles are practically one besides the other, which would imply a possible collision.

\section{Empirical data sets for the DL-based detection of oncoming vehicles}\label{database}

\begin{figure}[t]
	\centering
	\begin{subfigure}[h]{0.45\textwidth}
		\includegraphics[width=\textwidth,height=5cm]{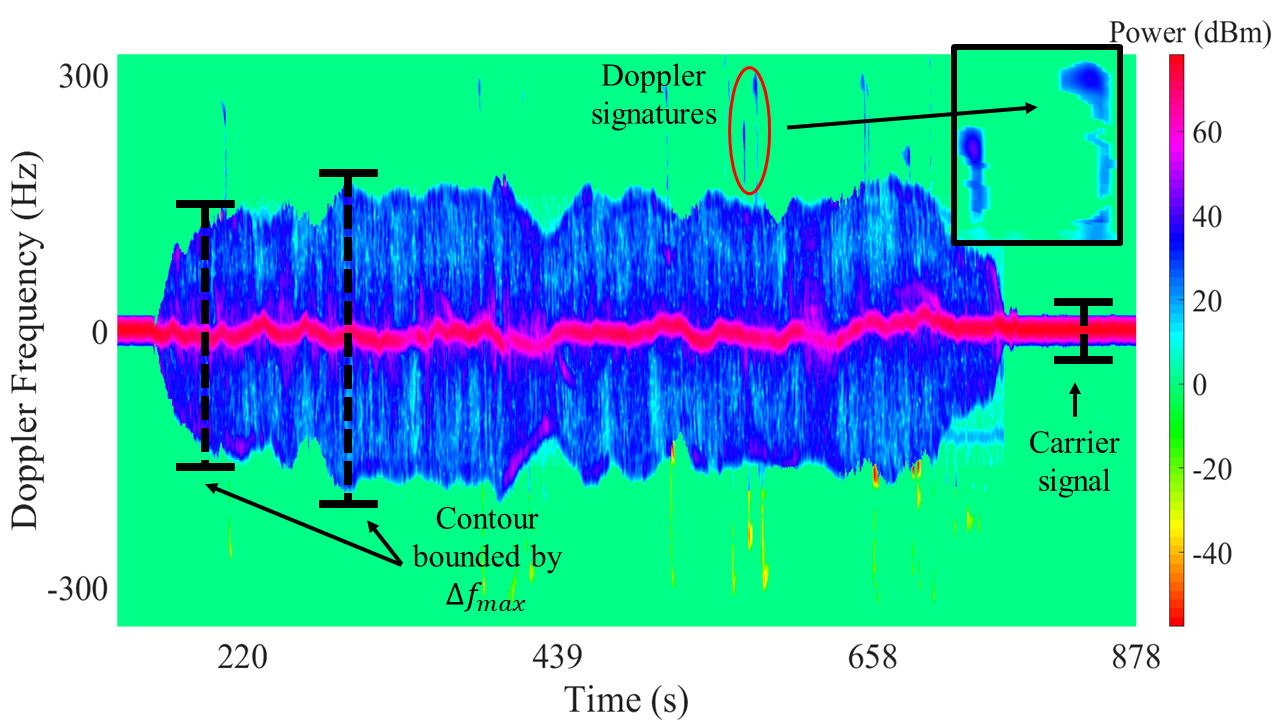}
		\caption{Spectrogram pre-processed using the data set of signals at 760~MHz.}
		\label{FIG:760clean}
	\end{subfigure}
	\hfill
	\begin{subfigure}[h]{0.45\textwidth}
		\includegraphics[width=\textwidth,height=5cm]{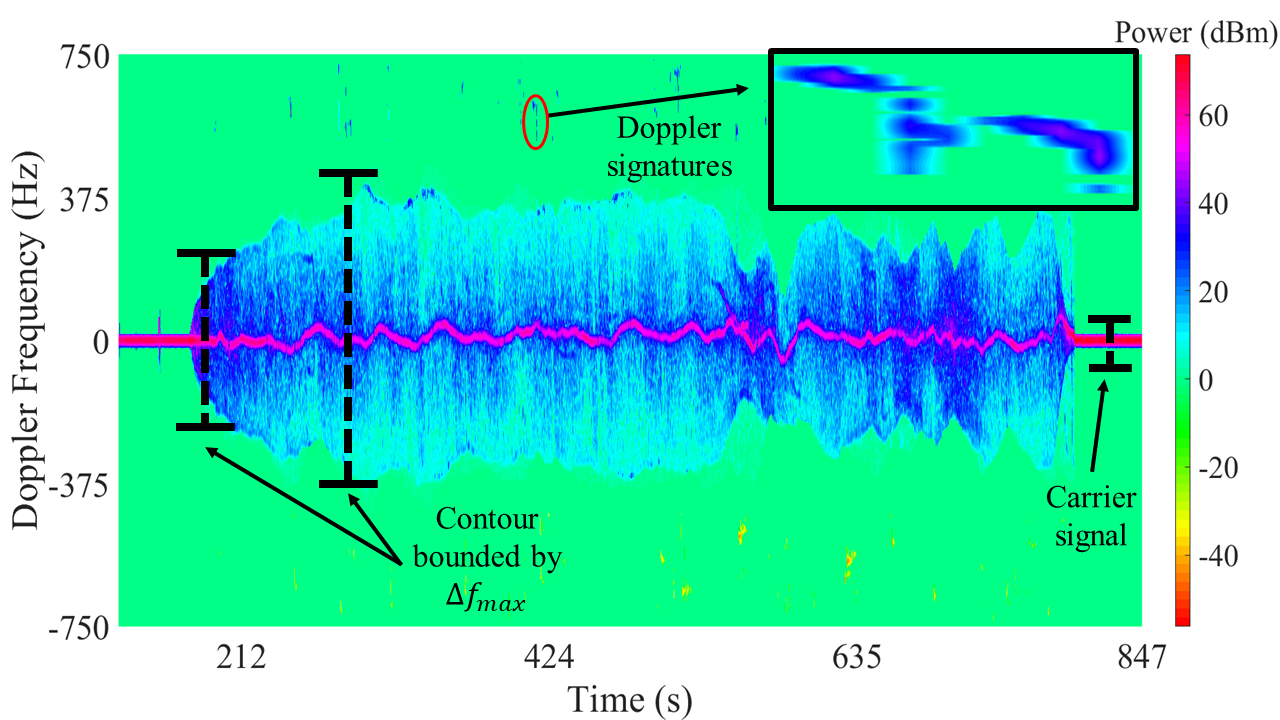}
		\caption{Spectrogram pre-processed using the data set of signals at 2.5~GHz.}
		\label{FIG:2500clean}
	\end{subfigure}
	\caption{Pre-processed Doppler spectrograms.}
	\label{FIG:ProcessedSpectrograms}
\end{figure}

The automatic classification and identification of the first part of the Doppler signature of an oncoming vehicle is addressed in this paper by applying DL tools. To apply these tools, it is necessary to denoise and correct the slow frequency drift of the measured Doppler spectra resulting from temperature variations of the measurement equipment. We followed the methods described in \cite{Gomez-Vega2022} to correct such issues and to synchronize the obtained data sets with the corresponding video files for visual inspection. Figure~\ref{FIG:ProcessedSpectrograms} shows two examples of the pre-processed (sanitized) Doppler spectrograms recorded at 760~MHz and 2.5~GHz. The LOS components of the received signal can be identified by the line in magenta, whereas the frequency band containing NLOS power received via reflections from static objects forms the contour shown in blue/light-blue. The length of both spectrograms spans only one half of the measurement circuit between points A and B. The absence of NLOS power at the beginning and the end of these spectrograms is due to the fact that the $T_X$ and $R_X$  made full stops for a few minutes after completing each trip between points A and B. This was done for the recording of short control data sequences.

Each data set collected at 760~MHz and 2.5~GHz is comprised of four spectrograms like those shown in Fig.~\ref{FIG:ProcessedSpectrograms}. The data set at 760~MHz is formed by a sequence of 6,553 Doppler spectra, for a total of 6,553$\times$1001 = 6,559,553 time-frequency (TF) samples. On the other hand, the data set at 2.5~GHz comprises a sequence of 6,047 spectra, for a total of 6,047$\times$1001 = 6,053,047 TF samples. It is convenient to reduce the bulk of data that needs to be processes in order to speed-up the execution time of the DL algorithms. From our analysis of the principal components of the recorded spectra, the data set at 760~MHz reveals that the meaningful TF samples are within the interval of $\pm$600~Hz. This allows to reduce the number of frequency samples per measured spectrum from 1001 to 425, resulting in a total of 2,785,025 TF samples to be processed by the DL algorithms. However, such a dimensionality reduction is not possible for the data set at 2.5~GHz,  because the Doppler dispersion at 2.5~GHz is more than three times higher than that observed at 760~MHz and the meaningful bandwidth of analysis is within $\pm1,000$~Hz for the signal transmitted at 2.5~GHz.

\section{Deep learning for forward-collision avoidance} \label{DL}
The use of data science for vehicular applications has been widely explored. 
One of the techniques that has raised most interest is DL \cite{Sarker2021}. This mathematical model aims to learn certain features of a data set by applying multiple layers of artificial neural networks. The typical architecture of a DL-based classification algorithm is comprised of a data training stage, a testing stage, and a third stage for the interpretation or validation of the algorithm. One of the advantages by which DL surpasses other techniques is that the extraction of the principal features of the data is done automatically.
Furthermore, once DL models learn to recognize the principal features through the training stage, they can use them to classify new data. This reduces the computational cost once the network is inferred \cite{Lu2020}. 
However, the accuracy in this case directly depends on the amount of data entered into the networks \cite{Sarker2021}. Therefore, it is important to have a large data set that allows algorithms to improve their performance.

\subsection{Long short-term memory network} \label{LSTM}
\begin{figure}[t]
	\centering
	\includegraphics[width=0.45\textwidth]{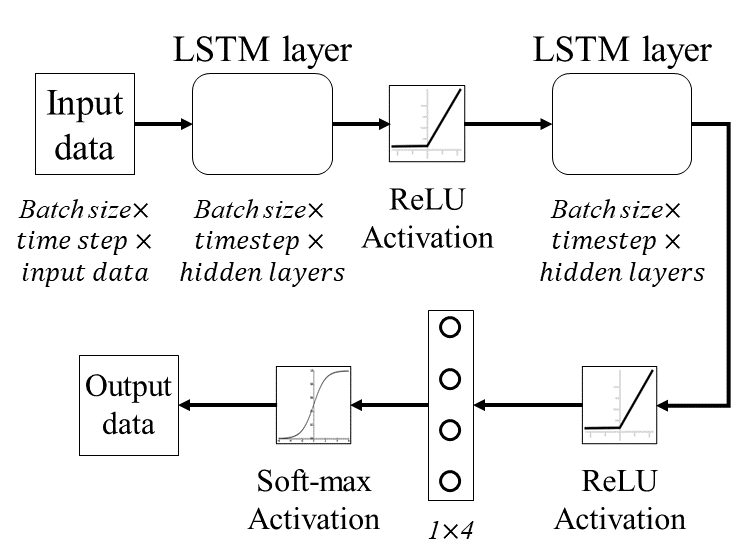}
	\caption{Neural network architecture using two layers of LSTM for feature extraction and a dense layer for classification.}
	\label{FIG:LSTM}
\end{figure}

In the analysis of sequential data, recurrent neural networks (RNN) have been widely used \cite{Liu2020}. 
However, in systems where the data sequences are very long, these networks present issues with the vanishing gradient. This makes it difficult for the network to have "memory" of the previous data. A variant of RNN known as an LSTM network has become popular to compensate for these practical issues \cite{LinArt2022}. 
We use two LSTM layers for the design of our classification system. The architecture of our LSTM network is shown in Fig.~\ref{FIG:LSTM}. The input data must have a three-dimensional representation corresponding to the batch size, the time step, and the number of features of the data set. The time step was set equal to one since we considered taking the smallest value of the observation window. This is because response time is essential to avoid possible collisions. Taking a larger window size would reduce the response time of our collision warning system. On the other hand, the number of features corresponds to the data registered in each frequency spectrum and depends on the frequency span of the measurement. The network parameters were chosen as follows: 
epoch = 800, 
batch size (training) = 5227; 
batch size (test) = 1326; 
time step = 1; 
hidden nodes = 100; 
features = 425-1001. Finally, the network uses a dense layer at the output to classify the samples into any of the four classes that were defined in Section \ref{database}.

\subsection{Convolutional neural network} \label{CNN}
\begin{figure}[t]
	\centering
	\includegraphics[width=0.45\textwidth]{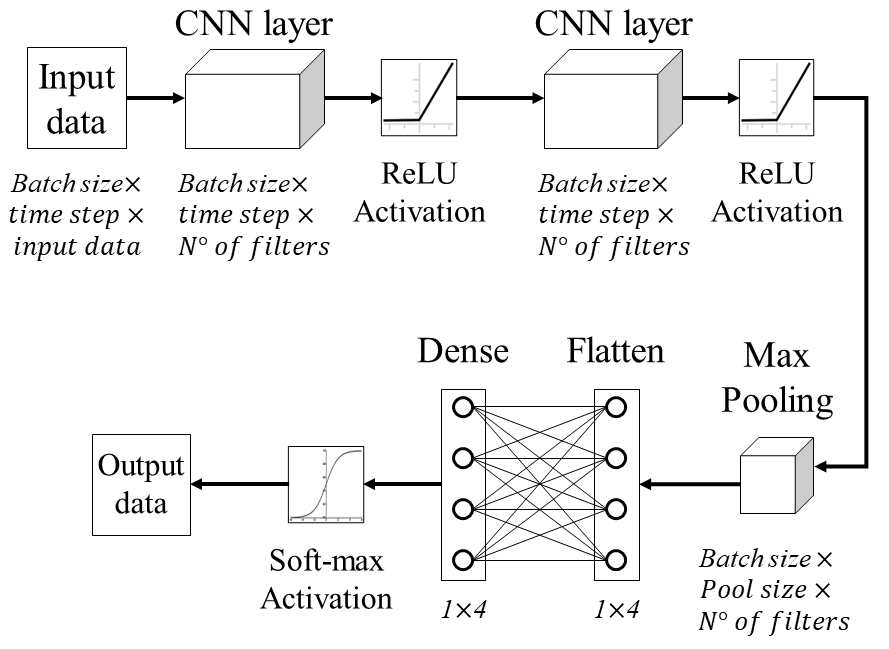}
	\caption{Neural network architecture using two layers of CNN for feature extraction, a max pooling layer for data reduction and a dense layer for classification.}
	\label{FIG:CNN}
\end{figure}

The measured spectrograms are a long data sequence that can be challenging for RNNs to analyze. Hence, to circumvent this issue, we implemented a different DL technique. CNN are popular because their architecture allows them to learn directly from input data without the need for a human-supervised feature extraction stage. Furthermore, this model can reduce the dimensionality of the data and maintain only the principal features. 
In CNN, a series of digital filters are used to compute the convolution of the input data and build a feature map for the classification process. The elements of our CNN are shown in Fig. \ref{FIG:CNN}. Data is processed by two convolution layers with an activation function for each set of filters and a specified kernel size. The number of filters can change according to the application in which they are used and depends on the feature map that is desired to be obtained. However, the filters must be sufficient to extract the most relevant features without consuming many computational resources for their implementation. As in the case of the LSTM network, data is feed in a configuration that includes batch size, time step, and the number of features. The chosen batch size is the same as for the LSTM case. The time step oscillates because in a CNN, the information is reduced by the network and it is necessary to maintain at least a minimum size to carry out the operations. A minimum range between 5 and 10 samples in the window observation size was selected to minimize the response time of the system. The number of features in this case also depends on the frequency span of the experimentation. On the other hand, the reduction in data dimensionality is carried out by the pooling layers. In a max pooling layer, the maximum values of the feature map of the convolution layer are taken and the rest are discarded. Furthermore, the size of the pooling operation has to be less than the size of the feature map and is commonly reduced by a factor of two. The last layer of the architecture is a fully connected layer that is used to process the reduced data and order it in such a way that the dense layer can receive it. This is achieved through a flatten layer. The summary of the values used in the network configuration is 
as follows: 
epoch = 800; 
batch size (training) = 5227; 
batch size (test) = 1326; 
stride = 1; 
time step = (5-10); 
number of filters = (8-64); 
kernel size = 3; 
pool size = 2; 
features = (425-1001).

\subsection{Neural networks performance metrics}

We considered four metrics to evaluate and compare the performance of the LSTM and CNN algorithms: accuracy, precision, recall, and specificity \cite{Bisong2019}. These metrics are in a normalized range that goes from zero to one (0\% and 100\%, respectively). Accuracy refers to how close the classification result was to the true values analyzed. Precision indicates how well each class was classified. The recall is the proportion of correctly classified data versus incorrectly classified data. And specificity expresses how effectively the model fits into identifying the classes. These metrics are computed using the true positives (TP), true negatives (TN), false negatives (FN), and false positives (FP) values. A TP is the number of samples correctly classified for a class. The TN refers to when the model has succeeded in properly classifying the samples that do not belong to the class being identified. The FN are the misclassified samples of a class. FP are samples from other classes that were classified as part of a class to which they do not belong. The performance metrics are defined as follows:
\begin{align}
\label{acc}
Accuracy &= \frac{TP+TN}{TP+FP+FN+TN}
\\
\label{precision}
Precision &= \frac{TP}{TP+FP}
\\
\label{recall}
Recall &= \frac{TP}{TP+FN}
\\
\label{speci}
Specificity &= \frac{TN}{TN+FP}.
\end{align}
Furthermore, we use the $R_X$-operating characteristic (ROC) curve and the area under the ROC curve (AUC) which are obtained from the relationship between the rates of TP and FP values. This allows us to quantitatively estimate the performance of the model. An AUC close to 1 indicates a better capacity to distinguish classes. The results of all these metrics are shown in the next section.

\section{Results and discussion} \label{results}
Using the data sets obtained for the CW probe signals at 760~MHz and 2.5~GHz, we classified the samples through the LSTM and CNN algorithms by considering four distinct events: static $T_X$ and $R_X $ (no-activity), moving $T_X$ and $R_X$ and no oncoming vehicle (no-oncoming vehicle), moving $T_X$ and $R_X$ and an oncoming vehicle approaching the $R_X$ (vehicle approaching), moving $T_X$ and $R_X$ and an oncoming vehicle driving away from the $R_X$ (vehicle driving away). For the learning stage, 80\% of the randomly selected samples were used and the remaining 20\% were used for validation. 

The results of the performance parameters of the LSTM model using both data sets are shown in Table \ref{tbl3}. The accuracy obtained in the case of experiments at 760~MHz was 96.94\%. With the data obtained from the 2.5~GHz signal, the accuracy was 96.53\%. However, even though the accuracy values in both cases are high, the class of vehicle approaching is the most important and determines the efficiency of the system. We compare the precision and sensitivity of this class for both data sets. It was found that there is a higher percentage in the experiments with signals at 2.5~GHz. The 98.60\% precision means that a high rate of samples was correctly classified and the system can detect if a car is approaching. Therefore, in a collision avoidance system, the results using the 2.5~GHz data set have better reliability.

\begin{table*}[t]
	\caption{LSTM performance evaluation.}
	\label{tbl3}
	\begin{center}
		\begin{tabular}{|c|c|c|c|c|c|c|c|c|}
			\hline
			Metric & \multicolumn{2}{c|}{\textbf{No activity}} & \multicolumn{2}{c|}{\textbf{No oncoming vehicle}} & \multicolumn{2}{c|}{\textbf{Vehicle approaching}} & \multicolumn{2}{c|}{\textbf{Vehicle driving away}}\\
			& 760~MHz & 2.5~GHz & 760~MHz & 2.5~GHz & 760~MHz& 2.5~GHz & 760~MHz & 2.5~GHz \\
			\hline
			Precision & 97.60\% & 98.70\% & 98.00\% & 96.40\% & \textbf{88.50\%} & \textbf{98.60\%} & 90.00\% & 98.00\% \\
			Recall & 99.60\% & 98.70\% & 97.80\% & 99.10\% & 75.40\% & 90.80\% & 95.70\% & 93.60\% \\
			Specificity & 99.41\% & 99.68\% & 95.53\% & 95.31\% & 99.51\% & 99.80\% & 99.15\% & 99.70\% \\
			AUC & 100\% & 99.00\% & 97.00\% & 97.00\% & 87.00\% & 95.00\% & 97.00\% & 97.00\% \\
			\hline
		\end{tabular}
	\end{center}
\end{table*}

\begin{table*}[t]
	\caption{CNN performance evaluation.}
	\label{tbl4}
	\begin{center}
		\begin{tabular}{|c|c|c|c|c|c|c|c|c|}
			\hline
			\textbf{Metric} & \multicolumn{2}{c|}{\textbf{No activity}} & \multicolumn{2}{c|}{\textbf{No oncoming vehicle}} & \multicolumn{2}{c|}{\textbf{Vehicle approaching}} & \multicolumn{2}{c|}{\textbf{Vehicle driving away}}\\
			& 760~MHz & 2.5~GHz & 760~MHz & 2.5~GHz & 760~MHz & 2.5~GHz & 760~MHz & 2.5~GHz \\
			\hline
			Precision & 96.78\% & 97.70\% & 96.10\% & 92.90\% & \textbf{68.40\%} & \textbf{94.50\%} & 78.70\% & 97.90\% \\
			Recall & 97.50\% & 97.70\% & 94.80\% & 97.90\% & 63.90\% & 79.10\% & 90.40\% & 91.10\% \\
			Specificity & 99.18\% & 99.46\% & 91.11\% & 90.58\% & 98.48\% & 99.31\% & 98.00\% & 99.70\% \\
			AUC & 98.00\% & 99.00\% & 93.00\% & 94.00\% & 81.00\% & 89.00\% & 94.00\% & 95.00\% \\
			\hline
		\end{tabular}
	\end{center}
\end{table*}

On the other hand, for the CNN classification model, the accuracy of the system using the signals captured at 760~MHz was 93.68\%. Using the data set obtained with the 2.5~GHz signals, the algorithm obtained an accuracy of 94.61\%. The results of the performance parameters computed with both data sets are shown in Table \ref{tbl4}. However, the precision in the detection of oncoming vehicle events is higher in the system when using the 2.5~GHz data set. Therefore, as in the case of the LSTM model, the system responds better when using a probe signal with a higher frequency.

Table \ref{tblAUC} shows the classification performance results for each model using the different data sets. As can be seen, the calculated AUC values were close to 1 for most classes. This indicates that the evaluated models had a high performance during their execution. According to these results, the DL LSTM algorithm obtained the highest classification performance in the event of approaching vehicles when using the data set containing the signals captured at 2.5~GHz. This model is known to have difficulties in retaining relevant information during the training stage. However, even though our data set contains several features, the classification model has satisfactory performance. In the case of the models that used the data set containing the signals captured at 760~MHz, it is shown that the model had the lowest performance in classifying the approaching vehicles. This may be because, during the pre-processing stage of the Doppler signatures, some of its principal features faded. Therefore, it is necessary to consider that the pre-processing stage has to be adapted if the carrier frequency is in another range to adequately preserve the features of their Doppler signatures.

\begin{table}
	\caption{Area under the ROC curve (AUC).}
	\label{tblAUC}
	\begin{center}
		\begin{tabular}{|c|c|c|c|c|}
			\hline
			\textbf{Dataset} & \multicolumn{2}{c|}{\textbf{760~MHz}} & \multicolumn{2}{c|}{\textbf{2.5~GHz}}\\
			DL model & LSTM & CNN & LSTM & CNN\\
			\hline
			AUC no activity & 1.0 & 0.98 & 0.99 & 0.99\\
			AUC no oncoming vehicle & 0.97 & 0.93 & 0.97 & 0.94\\
			AUC vehicle approaching & 0.87 & 0.81 & 0.95 & 0.89\\
			AUC vehicle driving away & 0.97 & 0.94 & 0.97 & 0.95\\
			\hline
		\end{tabular}
	\end{center}
\end{table}

\begin{figure}[t]
	\centering
	\includegraphics[width=0.5\textwidth, height= 5.5cm]{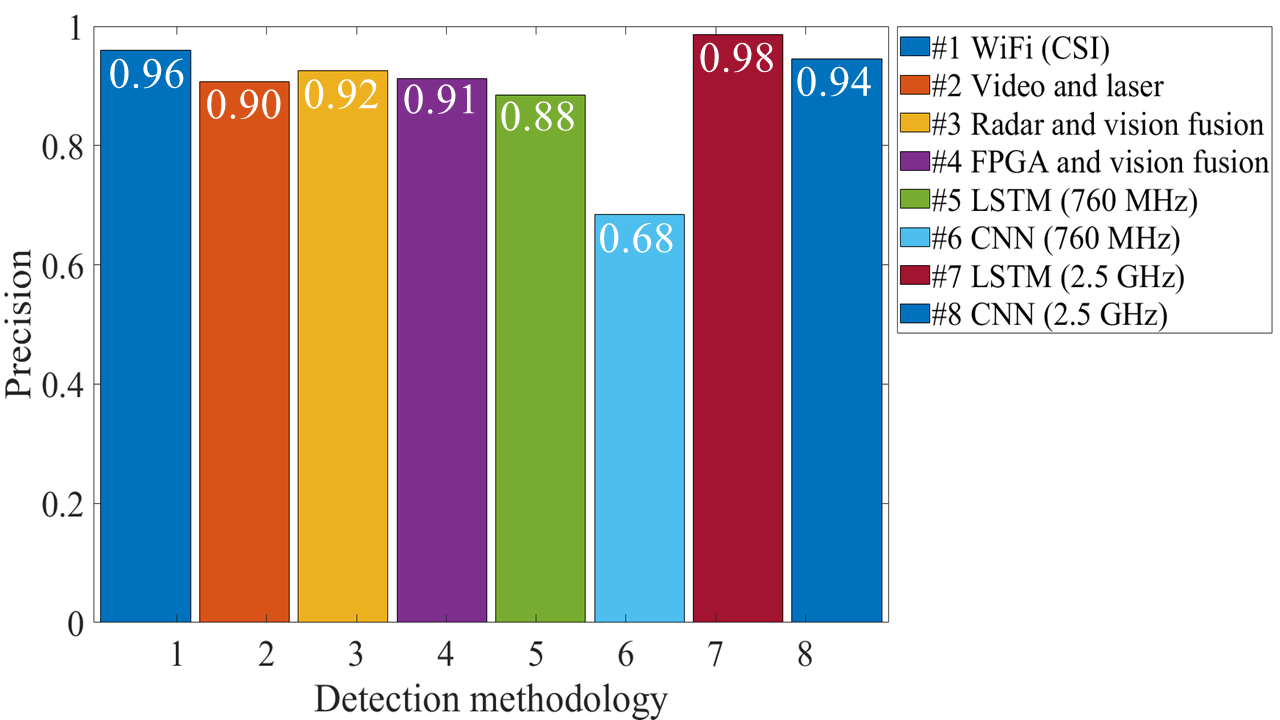}
	\caption{Comparison of precision achieved by different vehicle detection systems.}
	\label{FIG:results}
\end{figure}

Our system detects oncoming vehicles with a precision that can be compared to some of the published works on vehicle detection and recognition. Figure \ref{FIG:results} shows a comparison between the precision percentages achieved by the different configurations of our platform and the methodologies used in other works. For example, in \cite{Won2017} (methodology \#1) they used the WiFi channel state information to detect vehicles with a methodology based on machine learning and a support vector machine algorithm. Our system uses a DL approach and the accuracy achieved was, in some cases, higher. In \cite{Zhang2018a} (methodology \#2) a fusion between a vision system and the information from a laser sensor was used to detect vehicles at night. Therefore, the implementation requires specialized equipment which makes it expensive. In our case, we use general-purpose equipment to perform vehicle sensing under different environmental conditions. The authors in \cite{Jiang2021} (methodology \#3) proposes an innovative fusion of a radar and a vision system to assist autonomous cars. The resolution that radar systems can achieve is very practical for vehicle maneuver detection tasks. However, this requires specialized equipment and cannot be easily integrated into VCS. In the case of \cite{Chen2020} (methodology \#4), a panoramic camera and a DL approach were used, implementing a deep CNN network architecture. The principal disadvantage is that these systems must always maintain a direct LOS with the car, which cannot always be guaranteed. Our methodologies (\#5-\#8) work using the multi-path effect so they do not require a direct LOS and oncoming vehicles can be detected with a high precision rate.

Another critical component in this class of systems is the time it takes for vehicles to be warned of oncoming vehicles. Our DL models take this into account as part of their architecture. Both DL algorithms are trained with a sequential data set on which the warning signal can be emitted once the first Doppler signature of an oncoming vehicle is captured. In the case of the 760~MHz configuration, drivers could be alerted 0.92~s on average before intersecting an oncoming vehicle, with a standard deviation of 0.75~s. Furthermore, the maximum time for the alert was 4.4s before a possible collision. On the other hand, with a configuration using the 2.5~GHz signal, the average detection time was 0.94s with a standard deviation of 1.04~s. Moreover, the maximum detection time before the intersection was 9.2~s. According to several investigations, on average, people need between 390 and 600~milliseconds to detect and react to hazards on the road \cite{Wolfe2019}. Therefore, both systems based on a decentralized approach fused with DL classification algorithms are capable of warning drivers with a sufficient time window for decision-making.


\section{Conclusions}
\label{conclusion}
In this article, we experimentally validate the feasibility of a forward-collision avoidance system based on a decentralized RS approach and the analysis of the Doppler signatures of an oncoming vehicle. To compare and determine the highest accuracy that our system can achieve, the performance of both DL algorithms was compared using two different data sets. For both data sets, 4 different classes of activities were labeled to test the robustness of the system against different events. After several evaluations and tests by the classification algorithms, it was found that the highest accuracy of the system is achieved with the data set recorded at 2.5~GHz. Furthermore, with the LSTM algorithm, it is possible to detect 98.6\% of approaching vehicles with a response time for the driver of up to 9.6~s. This demonstrates that the decentralized RS approach described here is a feasible alternative for forward-collision warning systems based on the JRSAC paradigm.


\section*{Acknowledgments}
\noindent This work has been partially supported by Projects FONDECYT Regular 1211132 and STIC-AMSUD  AMSUD220026.

\bibliographystyle{ieeetr}
\bibliography{IEEEabrv,Vehicular2023}

\begin{IEEEbiography}[{\includegraphics[width=1in,height=1.25in,clip,keepaspectratio]{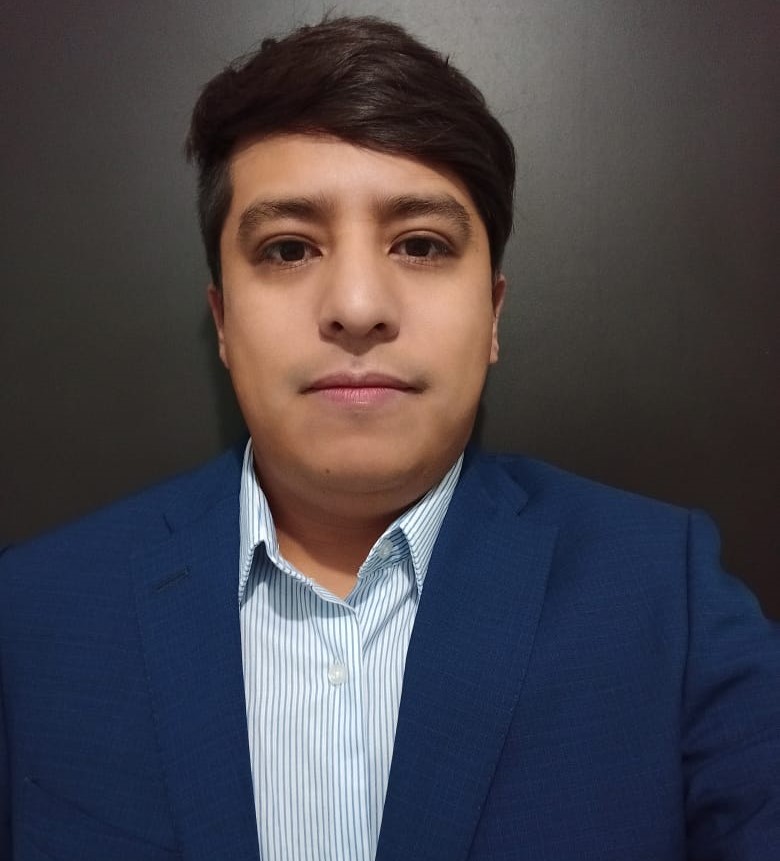}}]{Jorge C\'ardenas}
(Student Member) received the 
M.S. degree in applied science from the Optical Communications Research Institute, M\'exico, in 2018. He is currently pursuing a Ph.D. degree in engineering sciences from the Autonomous University of San Luis Potos\'i. He is a member of the IEEE Communications Society. His current interests include data science, digital processing of radio frequency signals, radio sensing, and wireless perception.\end{IEEEbiography}

\begin{IEEEbiography}[{\includegraphics[width=1in,height=1.25in,clip,keepaspectratio]{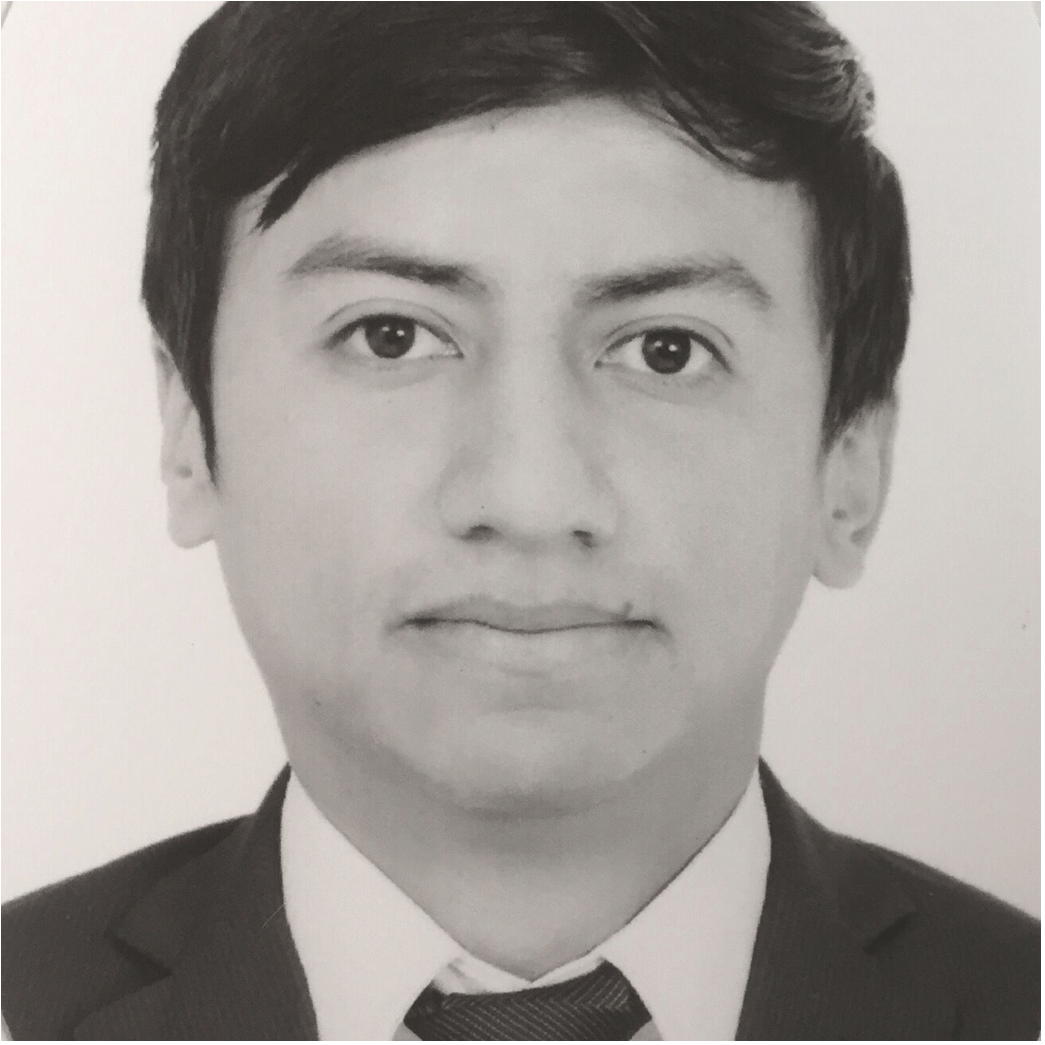}}]{Omar Contreras-Ponce}
received the 
M.S. degree in electronics in 2023, from the Universidad Autónoma de San Luis Potos\'i, M\'exico. He is currently a member of the radio frequency manufacture department at Robert Bosch Hermosillo Plant. His current research interests include electromagnetic theory for communications, antenna design, phase antenna arrays, and beam-forming transmission techniques.\end{IEEEbiography}

\begin{IEEEbiography}[{\includegraphics[width=1in,height=1.25in,clip,keepaspectratio]{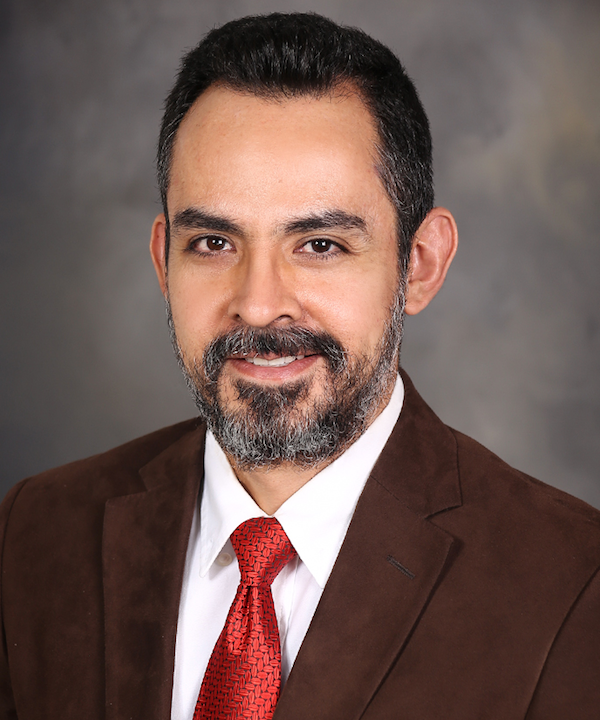}}]{Carlos A. Guti\'errez}
(Senior Member) received the 
Ph.D. degree in mobile communication systems from the University of Agder, Norway, in 2009. From 2009 to 2011, he was with the School of Engineering, Universidad Panamericana, Aguascalientes, M\'exico. Since January 2012, he has been with the Faculty of Science, Universidad Aut\'onoma de San Luis Potos\'i, Mexico. His research interests include modeling, simulation, and measurement of wireless channels; antenna design; vehicular communications; and wireless perception systems for vehicular applications and human activity recognition. 
\end{IEEEbiography}

\begin{IEEEbiography}[{\includegraphics[width=1in,height=1.25in,clip,keepaspectratio]{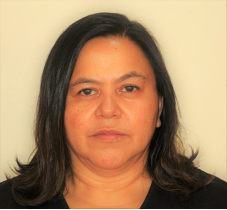}}]{Ruth Aguilar-Ponce}
(Member) received the 
Ph.D. degree from the University of Louisiana at Lafayette 
in 2007. 
She is currently an Assistant Professor with the Autonomous University of San Luis Potos\'i, M\'exico. Her research interests include sensor networks, video processing, signal and image processing, and embedded system design.\end{IEEEbiography}

\enlargethispage{-90ex}
\newpage

\begin{IEEEbiography}[{\includegraphics[width=1in,height=1.25in,clip,keepaspectratio]{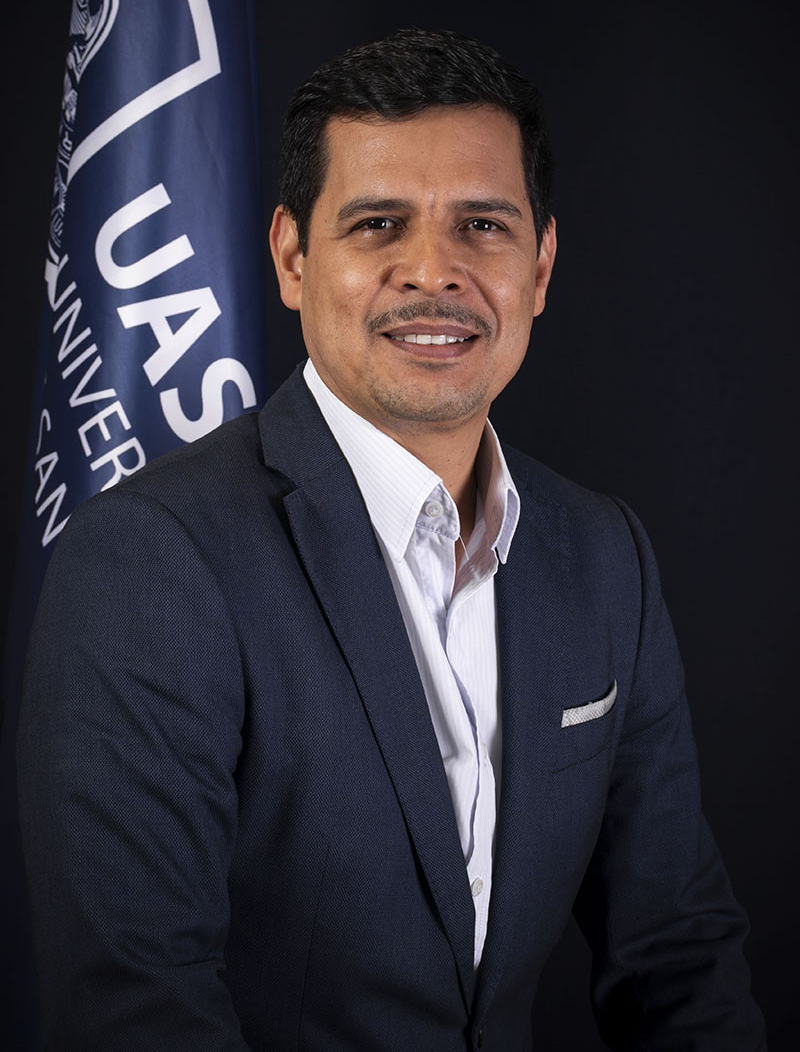}}]{Francisco R. Castillo-Soria}
(Member) received the 
degree of Doctor of Science in Electronics and Telecommunications by the Center for Scientific Research and Higher Education of Ensenada (CICESE) in 2015 in Ensenada B.C. Mexico. Since 2017 he is a Professor-Researcher at the Faculty of Sciences of the UASLP, M\'exico. His main lines of research are MIMO wireless communications, signal processing, spatial/index modulation, and modeling of multi-user MIMO-OFDM systems. 
\end{IEEEbiography}

\begin{IEEEbiography}[{\includegraphics[width=1in,height=1.25in,clip,keepaspectratio]{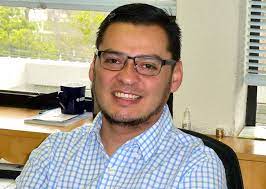}}]{Cesar A. Azurdia-Meza}
(Member) received 
a Ph.D. in Electronics and Radio Engineering from the University of Kyung Hee, Republic of South Korea, in 2013. He joined the Department of Electrical Engineering at the University of Chile as an Assistant Professor in August 2013. 
His research interests include topics such as the Nyquist ISI criteria, OFDM-based systems, SC-FDMA, visible light communication systems, in-vehicle communications, 5G enabling technologies and enabling technologies, as well as techniques signal processing for communication systems.
\end{IEEEbiography}

\enlargethispage{-90ex}

\end{document}